\documentclass[11pt]{article}

\usepackage[final]{acl}

\usepackage{times}
\usepackage{latexsym}

\usepackage[T1]{fontenc}

\usepackage[utf8]{inputenc}

\usepackage{microtype}

\usepackage{inconsolata}

\usepackage{graphicx}

\usepackage[table,xcdraw]{xcolor}
\usepackage{arydshln}
\usepackage{booktabs}
\usepackage{amsmath}
\usepackage{multirow}

\title{IterCOMP: Reasoning-aware Adaptive Prompt Compression \\ for Multi-hop Question Answering}

\author{
  \begin{large} Jungmin Yun\textsuperscript{1} \end{large} \textmd{and} \begin{large} Youngbin Kim\textsuperscript{1, 2} \end{large}\\\\
  \begin{normalsize}\textsuperscript{1}Department of Artificial Intelligence, Chung-Ang University \end{normalsize} \\
  \begin{normalsize}\textsuperscript{2}Graduate School of Advanced Imaging Sciences, Multimedia and Film, Chung-Ang University \end{normalsize} \\
  \begin{normalsize} \texttt{\{cocoro357, ybkim85\}@cau.ac.kr} \end{normalsize}
}

\begin{document}
\maketitle
\begin{abstract}
Multi-hop question answering requires complex reasoning across multiple evidence segments, which often overwhelms retrieval-augmented generation systems with lengthy and noisy contexts, thereby undermining both efficiency and accuracy. While existing prompt compression methods attempt to address this issue, they are typically designed for single-turn queries and fail to capture interdependent reasoning steps. We propose IterCOMP, a unified, training-free prompt compression framework that incorporates multi-hop reasoning within an iterative compression loop. IterCOMP decomposes documents into evidence segments, evaluates question answerability, and generates targeted follow-up questions to iteratively integrate essential evidence, producing a compact, reasoning-oriented prompt. Experiments on MusiQue, 2WikiMultiHopQA, and HotpotQA demonstrate that IterCOMP achieves substantial improvements in Exact Match and F1 scores while reducing the token budget, outperforming existing baselines and exhibiting robustness as reasoning complexity increases.
\end{abstract}

\section{Introduction}

Multi-hop question answering (QA) requires reasoning over multiple pieces of evidence to derive the correct answer. Large Language Models (LLMs) have significantly advanced QA performance by enhancing complex query understanding and information integration. Nevertheless, their reliance on static pre-training data imposes inherent limitations on knowledge coverage. Retrieval-Augmented Generation (RAG)~\cite{lewis2020retrieval} mitigates this issue by incorporating dynamic, up-to-date information from external sources, thereby enabling LLMs to generate more diverse, accurate, and contextually grounded responses~\cite{gao2023retrieval}.

\begin{figure}[t!]
  \centering
  \includegraphics[width=0.95\linewidth]{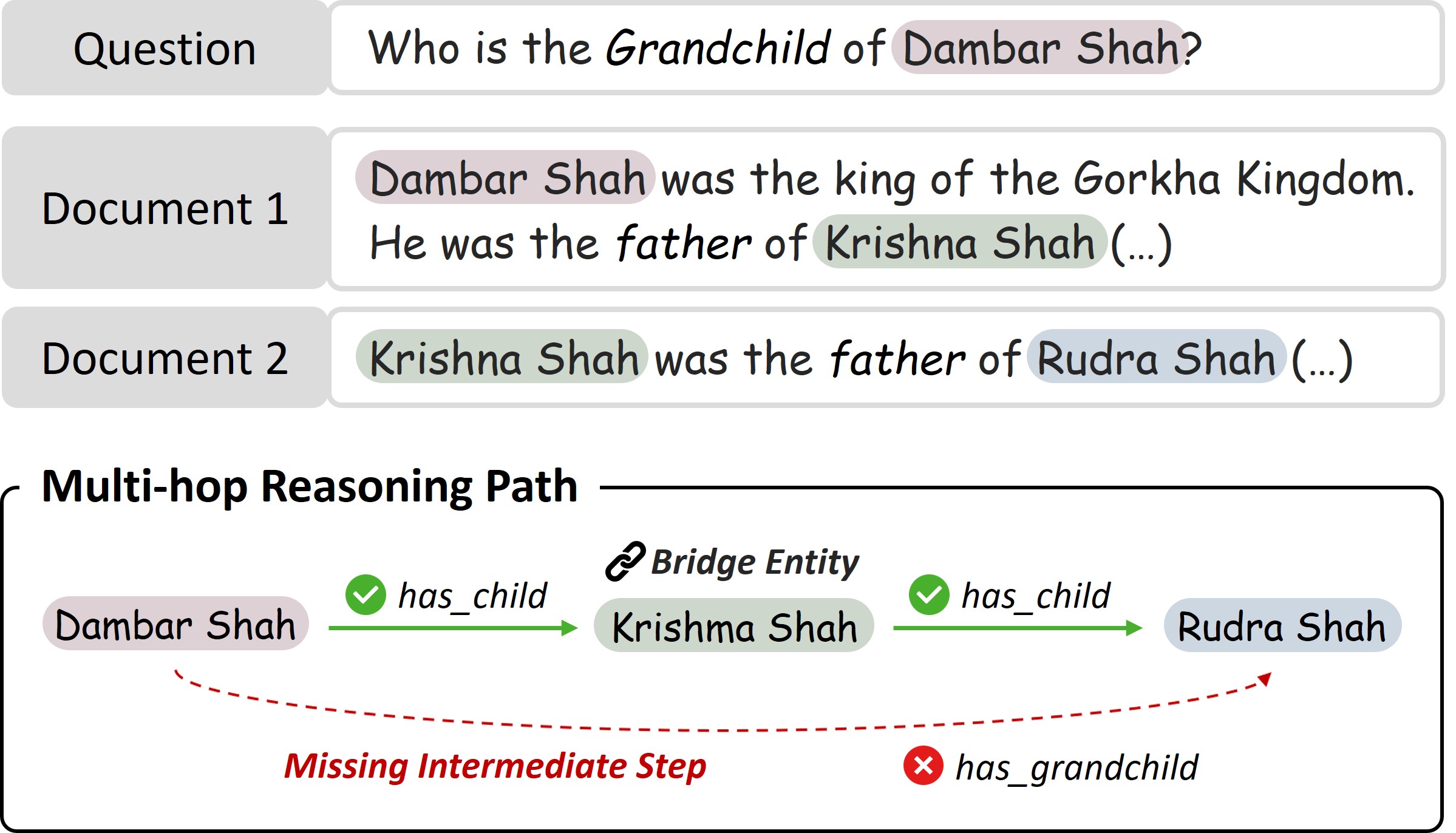}
  \caption{Example of multi-hop reasoning, where answering the question requires integrating evidence from multiple documents through implicit intermediate clues absent from the initial question.}
  \label{teasure_fig}
\end{figure}

However, RAG systems face challenges in both efficiency and effectiveness due to the long input sequences induced by retrieved documents. Efficiency declines as input length scales linearly with the number of retrieved documents, leading to higher inference latency and increased computational overhead~\cite{li-etal-2025-brief, xu2024recomp}. For API-based commercial LLMs, longer inputs also result in elevated operational costs~\cite{choi-etal-2024-reading}. Effectiveness is similarly constrained, as lengthy inputs often include irrelevant content that distracts the model and hinders accurate reasoning~\cite{shi2023large}. Furthermore, LLMs exhibit positional biases, such as the lost-in-the-middle phenomenon~\cite{liu-etal-2024-lost}. These challenges are especially pronounced in multi-hop QA, which requires linking multiple pieces of evidence through sequential and interdependent reasoning steps to derive the final answer~\cite{tang2024multihop}.

Recently, various prompt compression techniques have been proposed to reduce contextual overhead by eliminating less salient content or condensing information into compact representations~\cite{pan-etal-2024-llmlingua, jiang-etal-2024-longllmlingua, mu2023learning}. However, effective compression requires more than simply reducing input length; it fundamentally depends on selectively retaining query-relevant information query-relevant while simultaneously removing irrelevant or distracting content~\cite{cao-etal-2024-retaining}. 

In this context, query-focused compression methods enhance contextual relevance by condensing prompts to retain content pertinent to a given query~\cite{choi-etal-2024-reading, liskavets2025prompt, hwang-etal-2025-exit}. However, their predominant reliance on a single-query paradigm, which primarily leverages surface-level query-document relevance, limits their effectiveness in complex scenarios such as multi-hop QA~\cite{trivedi-etal-2023-interleaving, press-etal-2023-measuring, shao-etal-2023-enhancing}. As illustrated in Figure~\ref{teasure_fig}, these tasks require sequential, interdependent reasoning across multiple documents and often depend on implicit intermediate clues absent from the initial query~\cite{schnitzler2024morehopqa, geva2021did, trivedi-etal-2022-musique, ho2020constructing}. These limitations become more pronounced when queries consist of multiple subcomponents, whose supporting evidence is dispersed across different documents~\cite{levy-etal-2025-documents}. Consequently, single-query approaches often fail to capture inter-document dependencies or to reason effectively over linked information, leading to information loss and incomplete contextual understanding that is inadequate for synthesizing multi-source answers.

To this end, we propose IterCOMP, a unified prompt compression framework based on iterative refinement that explicitly incorporates multi-hop reasoning into the compression loop. IterCOMP strategically integrates the reasoning capabilities of LLMs into the compression process to directly address the nuanced evidential demands of complex queries. Specifically, the framework filters relevant evidence segments and employs an LLM to assess whether they are sufficient to answer the question. When the evidence is insufficient, the LLM identifies the informational gap and formulates a targeted follow-up question to bridge it. This initiates a cycle in which new evidence is evaluated against the evolving reasoning path, with only critical segments retained and irrelevant content discarded. Through progressive accumulation and distillation, IterCOMP incrementally constructs a concise yet comprehensive prompt that enables the original multi-hop question to be answered using focused, essential information. Extensive experiments demonstrate that IterCOMP substantially improves both QA performance and efficiency, highlighting the effectiveness of integrating deep reasoning into prompt compression.

\section{Related Work}

\subsection{Prompt Compression}

\subsubsection{Soft Prompt Compression}
Soft prompt compression encodes the original prompt into continuous vector representations~\cite{gecontext, cheng2024xrag}. AutoCompressor~\cite{chevalier-etal-2023-adapting} segments long prompts into multiple parts, compresses each segment into a soft prompt representation, and concatenates them to form the final prompt. GIST~\cite{mu2023learning} generates prefix-style soft prompts for instructional inputs. These methods enable parameter-efficient adaptation of pretrained models while preserving the high-level semantics of the original input. However, soft prompts are typically optimized for specific LLMs, which limits their transferability across different models~\cite{li-etal-2025-prompt}. This constraint poses a significant challenge in API-based environments, where direct access to model internals is restricted. Consequently, any update to the underlying LLM requires retraining the soft prompts, reducing their practicality in dynamic or cross-model deployment scenarios.

\subsubsection{Hard Prompt Compression}

Hard prompt compression reduces prompt length by preserving essential information through extractive or abstractive methods~\cite{li-etal-2025-prompt, chuang-etal-2024-learning, 10535182}. This approach improves computational efficiency and enhances the quality of generated outputs by eliminating redundant or uninformative content. 

\noindent \textbf{Query-Agnostic Compression.} Query-agnostic methods operate independently of the query, leveraging the statistical or structural properties of the prompt. Selective-Context~\cite{li-etal-2023-compressing} discards low-information tokens based on self-information metrics. LLMLingua~\cite{jiang-etal-2023-llmlingua} removes tokens with low perplexity, while LLMLingua-2~\cite{pan-etal-2024-llmlingua} formulates compression as a token classification task using knowledge distillation. Despite their broad applicability, these approaches overlook query-specific context. As a result, compressed prompts may retain irrelevant content or omit critical information, thereby degrading retrieval effectiveness and response relevance~\cite{cao-etal-2024-retaining}. 

\noindent \textbf{Query-Aware Compression.} Incorporating query information into the compression process is essential for retaining relevant and critical content for reasoning. LongLLMLingua~\cite{jiang-etal-2024-longllmlingua} employs a coarse-to-fine strategy, first estimating document-level importance via query-conditioned perplexity, followed by refining token selection. COMPACT~\cite{yoon-etal-2024-compact} compresses context by jointly analyzing previously selected content and newly introduced segments. RECOMP~\cite{xu2024recomp} performs sentence-level compression by measuring the similarity between query and sentence embeddings to identify key content. R2C~\cite{choi-etal-2024-reading} encodes the query alongside each context chunk, enabling the decoder to dynamically preserve relevant information.

Despite their advantages, existing query-aware methods typically assess relevance by matching a single query against individual documents or sentences. This one-to-one paradigm is insufficient for complex multi-hop QA, where a query often comprises multiple subcomponents that must be jointly resolved using evidence distributed across documents~\cite{tang2024multihop}. Such tasks require capturing inter-document dependencies and reasoning over linked information segments~\cite{trivedi-etal-2023-interleaving}. However, current methods struggle to model these multi-hop relationships, thereby limiting their effectiveness in synthesizing information from multiple sources~\cite{zhu-etal-2025-mitigating}.

\subsection{Self-Ask Mechanisms in LLMs}
Recent studies highlight the potential of LLMs to generate and respond to follow-up questions, thereby enhancing performance across various tasks. Follow-up questioning improves the coherence and informativeness of document generation~\cite{tix2024follow} and increases user satisfaction in conversational search by supporting deeper exploration~\cite{kim2024using}. A prominent approach is the self-ask method, in which LLMs generate and answer sub-questions to decompose complex queries, yielding substantial gains in compositional reasoning~\cite{press-etal-2023-measuring}. This line of work has been extended to interleave reasoning, retrieval, and self-reflection, allowing LLMs to dynamically control queries during problem solving~\cite{jiang-etal-2023-active, yao2023react, asai2024self}. Domain-specific adaptations demonstrate that iterative RAG can effectively address complex clinical scenarios~\cite{xiong2024improving}. In broader contexts, additional work explores anticipating follow-up questions in information search~\cite{wilcock-2024-anticipating}, refining outputs through targeted questioning~\cite{shridhar-etal-2024-art}, and augmenting LLMs with search engines to improve factuality~\cite{vu-etal-2024-freshllms}.

\section{Preliminary Analysis}

Our proposed approach is grounded in two key assumptions: (1) LLMs can reliably judge whether a question is answerable given a set of evidence; and (2) if the question is unanswerable, LLMs can identify the additional information required to derive the correct answer. To empirically validate these assumptions, we conduct two preliminary analyses guided by the following research questions: \textbf{\textit{RQ1: Can LLMs determine whether the provided evidence is sufficient to answer a given question? RQ2: If the evidence is insufficient, can LLMs identify the missing information needed to derive the correct answer?}}

\subsection{Settings}
To address these questions, we employ the MuSiQue dataset~\cite{trivedi-etal-2022-musique}, which consists of multi-hop QA pairs requiring two to four reasoning steps. For each hop length, we randomly sample 400 QA pairs. Following prior work~\cite{wang-etal-2025-llms}, we use \texttt{GPT-4o} to generate sub-question and sub-answer pairs based on the decomposition annotations provided in the dataset. Table~\ref{tab:analysis} presents examples of the prompts and resulting generated data.

Using the original question along with the generated sub-QA pairs, we conduct experiments with several LLMs, including \texttt{LLaMA-3.1-8B-Instruct}~\cite{dubey2024llama}, \texttt{Mistral-7B-Instruct}~\cite{jiang2023mistral7b}, \texttt{GPT-3.5-Turbo}~\cite{brown2020languagemodelsfewshotlearners}, and \texttt{GPT-4o}~\cite{achiam2023gpt}. Each model is evaluated on two tasks: (1) determining whether the provided evidence alone is sufficient to answer the original question, and (2) identifying the additional information required when the evidence is deemed insufficient.

\subsubsection{Answerability Judgment} To evaluate answerability judgment, we define two experimental conditions based on the completeness of the provided evidence:
\begin{itemize}
    \item \textit{Full}: The evidence includes sub-answers for all hops required to answer the original question (e.g., all three sub-answers in a 3-hop QA pair) → \textit{answerable}.
    \item \textit{Partial}: The evidence includes sub-answers only up to an intermediate hop (e.g., only the 1st and 2nd hops in a 3-hop QA pair) → \textit{unanswerable}.
\end{itemize}

\subsubsection{Missing Information Identification.} For missing information identification, we adopt a \textit{partial} evidence setting. Specifically, the sub-answers up to a given hop are concatenated and provided as evidence, while the sub-question at the subsequent hop is treated as the reference missing information that the model is required to identify. 

To evaluate model predictions, we employ two metrics. BLEU score measures surface-level similarity between the generated follow-up question and the reference sub-question by capturing n-gram overlap. Relatedness score (Rel.) is computed through binary classification using \texttt{GPT-4o}, assessing whether the predicted and reference questions are topically or conceptually aligned.

\subsection{Results}

\begin{table}[t!]
\resizebox{1.0\columnwidth}{!}{%
\begin{tabular}{l|ccc|c|ccc|c}
\toprule
                      & \multicolumn{4}{c|}{\textbf{\textit{Full}}}                                                       & \multicolumn{4}{c}{\textbf{\textit{Partial}}}                                                     \\ \cmidrule{2-5} \cmidrule{6-9}
 
\textbf{}             & \textbf{2-hop} & \textbf{3-hop} & \textbf{4-hop} & \textbf{\textit{Avg.}} & \textbf{2-hop} & \textbf{3-hop} &\textbf{ 4-hop} & \textbf{\textit{Avg.}} \\ \midrule
\texttt{Llama-3.1-8B} & 86.5          & 58.0          & 47.3          & 63.9         & 71.3          & 93.5          & 96.5          & 87.1         \\ 
\texttt{Mistral-7B}   & 92.0          & 80.3          & 78.5          & 83.6         & 71.8          & 95.5          & 97.6          & 88.3         \\ 
\texttt{GPT-3.5}      & 68.3          & 63.3         & 58.0             & 63.2         & 88.3          & 96.0          & 96.9          & 93.7         \\ 
\texttt{GPT-4o}       & 81.8          & 72.5          & 67.8          & 74.0         & 95.5          & 98.3          & 99.5          & 97.8         \\ \bottomrule
\end{tabular}
}
\caption{Results on answerability judgment for information sufficiency assessment.}
\label{Tab:1}
\end{table}

\begin{figure*}[t!]
  \centering
  \includegraphics[width=1.0\textwidth]{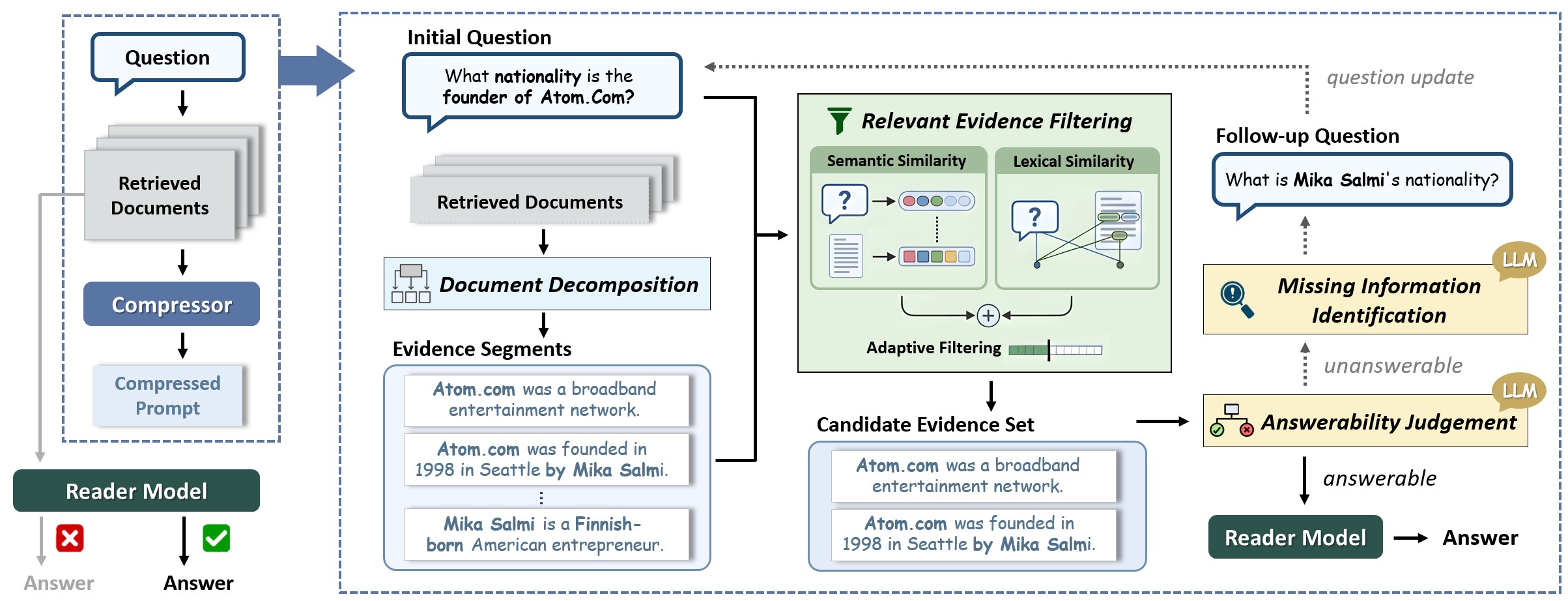}
  \caption{Overall pipeline of IterCOMP. Retrieved documents are decomposed into evidence segments and filtered via dual-aspect relevance scoring. An LLM then judges whether the candidate evidence set is sufficient to answer the question. If answerable, the evidence is finalized as the compressed prompt and passed to the reader model. If unanswerable, the LLM identifies missing information and generates a targeted follow-up question, which serves as a new query for the filtering stage to enable iterative evidence accumulation.}
\end{figure*}

\noindent
\textbf{Answerability Judgment.} Table~\ref{Tab:1} presents the results of the answerability judgment task across various LLMs under two conditions: \textit{Full}, where all required evidence (i.e., sub-answers for each reasoning step) is provided, and \textit{Partial}, where only a subset of evidence is available.

The models exhibit contrasting trends across the two conditions. In the \textit{Full} setting, accuracy consistently declines as the number of hops increases; for example, \texttt{Mistral-7B} drops from 92\% at 2-hop to 78.5\% at 4-hop. This indicates that, even when all relevant evidence is available, increasing hop length amplifies both the volume of evidence and the complexity of their interconnections. Accordingly, answerability judgment requires more than surface-level binary classification and instead demands advanced reasoning capabilities to integrate multiple evidence segments. In contrast, under the \textit{Partial} setting, all models show consistent improvements as hop length grows. High-performing models such as \texttt{GPT-4o} maintain strong performance with minimal variance even at longer hop lengths. For a more balanced evaluation, we also report overall F1 scores: \texttt{Llama-3.1-8B} (70.51), \texttt{Mistral-7B} (84.27), \texttt{GPT-3.5-Turbo} (73.12), and \texttt{GPT-4o} (83.57).

Although these scores reveal an asymmetry in detecting answerable and unanswerable cases, both misclassification types pose distinct risks. Misclassifying an unanswerable case as answerable causes the compression process to terminate prematurely, leaving the reasoning chain with insufficient clues and thus compromising the reliability of the final answer. Conversely, misclassifying an answerable case as unanswerable triggers unnecessary iterations; while this allows the reasoning chain to incorporate new evidence, prolonged iterations may introduce irrelevant noise that degrades compression effectiveness. Given the relative degradation observed in the \textit{Full} setting, we bound the maximum number of iterations to prevent excessive refinement. Building on this analysis, we design IterCOMP's iterative compression mechanism with an iteration budget calibrated to these tendencies, reliably using answerability as a core signal to guide evidence collection in complex multi-hop QA.

\noindent
\textbf{Missing Information Identification}. Table~\ref{Tab:2} presents the experimental results for the missing information identification task, which evaluates the quality of follow-up questions generated by LLMs when the initial evidence is insufficient to answer the original question. The analysis shows that \texttt{GPT-4o} achieves the best performance, with an average relatedness score of 90.6\% and an average BLEU score of 0.50. These results highlight \texttt{GPT-4o}'s strong capability to pinpoint missing information and generate contextually appropriate follow-up questions within multi-hop reasoning contexts. However, all models exhibit a consistent decline in both relatedness and BLEU scores as the number of hops increases. This trend reflects the growing difficulty of accurately identifying knowledge gaps and formulating effective follow-up questions as reasoning chains become longer and more complex. While LLMs clearly demonstrate the capability to generate meaningful follow-up questions, the observed performance degradation suggests that precisely locating and articulating missing information remains a challenging subtask in multi-hop QA.

\begin{table}[t!]
\resizebox{1.0\columnwidth}{!}{%
\begin{tabular}{l|cc|cc|cc|cc}
\toprule

                               & \multicolumn{2}{c|}{\textbf{2-hop}} & \multicolumn{2}{c|}{\textbf{3-hop}} & \multicolumn{2}{c|}{\textbf{4-hop}} & \multicolumn{2}{c}{\textit{\textbf{Avg.}}} \\ \cmidrule{2-3} \cmidrule{4-5} \cmidrule{6-7} \cmidrule{8-9}

\textbf{}                      & \textbf{BLEU}               & \textbf{Rel.}              & \textbf{BLEU}               & \textbf{Rel.}               & \textbf{BLEU}               & \textbf{Rel.}               & \textbf{BLEU}              & \textbf{Rel.}              \\ \midrule
\texttt{Llama-3.1-8B} & 0.40  & 81.3 & 0.31             & 75.8  & 0.27  & 70.3  & 0.30   & 73.8   \\ 
\texttt{Mistral-7B}   & 0.52     & 92.0    & 0.43    & 88.1     & 0.37 & 82.1  & 0.41  & 85.6  \\ 
\texttt{GPT-3.5}               & 0.44      & 88.5    & 0.37       & 87.8 & 0.30     & 78.7          & 0.35        & 85.4 \\ 
\texttt{GPT-4o}                & 0.61     & 97.9      & 0.58     & 91.9        & 0.41 & 87.2           & 0.50                      & 90.6  \\ \bottomrule
\end{tabular}
}
\caption{Results on missing information identification.}
\label{Tab:2}
\end{table}

\section{Methodology}

\subsection{Problem Formulation}

We define the task of prompt compression as follows. Given an initial question $q$ and a large corpus of retrieved documents $D=\{d_{1},\ldots,d_{N}\}$, the primary objective is to synthesize a compressed prompt $P_\text{comp}$. The prompt $P_\text{comp}$ consists of a concise yet information-sufficient set of evidence from $D$, satisfying the constraint $L(P_\text{comp}) \ll \sum_{i=1}^{N} L(d_i)$, where $L(\cdot)$ denotes the token length. The principal aim is to construct a $P_\text{comp}$ such that a downstream reader model $M$ can generate a high-fidelity output, $y = M(P_\text{comp}, q)$, while preserving essential reasoning information from corpus $D$. 

In multi-hop QA, where synthesizing information across multiple sources is indispensable, prompt compression plays a pivotal role. Its utility extends beyond filtering irrelevant content and selecting evidence that directly addresses the initial question $q$. Multi-hop reasoning frequently requires resolving latent intermediate questions not explicitly stated in $q$. Consequently, an effective compression strategy must support progressive evidence accumulation, enabling the model to incrementally integrate and build upon relevant information as the reasoning process unfolds. This capability is fundamental to maintaining logical coherence across complex inferential chains and mitigating performance degradation due to attentional dilution or contextual drift in long inputs.

\subsection{Proposed Framework}
\subsubsection{Document Decomposition} Each document $d_i \in D$ is segmented into a set of smaller units, termed \textit{evidence segments} and denoted by $E_{i}=\{e_{i,1}, \ldots, e_{i,R_i}\}$, where $R_i$ indicates the number of segments for $d_i$. Although segments can be defined at various levels of granularity, such as paragraphs or sentences, we adopt sentence-level decomposition in this work. A sentence serves as a basic unit that preserves the local semantic context of the original input while enabling effective compression~\cite{xu2024recomp, choi-etal-2024-reading}. These evidence segments, along with the initial question $q$, are then passed to the relevant evidence filtering module.

\subsubsection{Relevant Evidence Filtering} The relevant evidence filtering module systematically identifies and retains salient evidence segments by evaluating the relevance of each candidate segment $e_{i,j}$ with respect to a given question $q$. To establish a robust foundation for evidence selection, our approach combines semantic and lexical signals in a unified scoring scheme.

\textit{Semantic similarity} captures contextual alignment between the question $q$ and a candidate evidence segment $e_{i,j}$ to encode deeper semantic correspondences. We transform both $q$ and $e_{i,j}$ into dense vectors representations via a text encoder $E(\cdot)$. The semantic similarity score, $S_\text{sem}(q,e_{i,j})$, is then computed as the inner product between the corresponding embeddings:
\begin{equation}
    S_\text{sem}(q,e_{i,j}) = E(q)^{\top}E(e_{i,j}),
\end{equation}
\\where $E(q)$ and $E(e_{i,j})$ denote the embeddings of the question $q$ and the evidence segment $e_{i,j}$, respectively.

While semantic similarity effectively captures broader contextual relevance, it may overlook critical lexical cues. In multi-hop settings, sensitivity to precise keyword matches improves coverage and accuracy~\cite{zhangsirerag}. To complement this, we incorporate a \textit{lexical similarity} component that quantifies token-level relevance based on the importance of overlapping terms between $q$ and $e_{i,j}$. Following M3-Embedding~\cite{chen-etal-2024-m3}, we define the lexical importance weight of each token $t$ in the question $q$ as follows:
\begin{equation}
    w_t^q=\text{ReLU}({\mathbf{w}_\text{lex}}^\top E(t)),
\end{equation}
\\where $\mathbf{w}_\text{lex}$ is a projection vector that maps a contextualized token representation to a scalar importance score, and $E(t)$ denotes the contextualized embedding of token $t$. An identical operation is applied to compute $w^{e_{i,j}}_t$ for tokens in the evidence segment $e_{i,j}$. The lexical similarity score is then defined as:

\begin{equation}
    S_\text{lex}(q,e_{i,j})=\sum\limits_{t \in q \cap e_{i,j}} w^q_t \cdot w^{e_{i,j}}_t,
\end{equation}
\\which computes a weighted sparse inner product over co-occurring tokens, capturing their joint lexical salience under respective contexts.

For both the encoder and the projection vector $\mathbf{w}_\text{lex}$, we adopt the pretrained bge-m3 model~\cite{chen-etal-2024-m3} without any additional fine-tuning. All parameters remain frozen throughout the entire IterCOMP pipeline, in line with our training-free design principle.

Finally, we combine the semantic and lexical signals into a single \textit{dual-aspect relevance score}:

\begin{equation}
\begin{aligned}
S_\text{dual}(q,e_{i,j}) &= \lambda \cdot S_\text{sem}(q,e_{i,j}) \\
                         &\quad + (1-\lambda) \cdot S_\text{lex}(q,e_{i,j}),
\end{aligned}
\end{equation}
\\where the hyperparameter $\lambda \in [0,1]$ balances semantic and lexical relevance.

To prune less relevant segments, we adopt percentile-based filtering that adaptively selects evidence according to relative importance within the retrieved context. Let $\mathcal{S}=\{S_\text{dual}(q, e_{i,j})\mid\forall i,j\}$ denote the multiset of relevance scores for all candidates. The filtered candidate set  $E_\text{cand}$ is constructed by retaining segments whose scores exceed the $k$-th percentile threshold over the global score distribution: 
\begin{equation}
E_\text{cand}=\{e_{i,j}\:|\: S_\text{dual}(q,e_{i,j}) \geq \text{Percentile}(\mathcal{S},k)\}.
\end{equation}
The segments retained in $E_\text{cand}$ form the basis for constructing the final compressed prompt $P_{\text{comp}}$. During multi-hop inference, this adaptive filtering is applied at each reasoning step, enforcing a dynamic cutoff that suppresses noise propagation and guides the model to attend to salient evidence throughout the reasoning chain.

\begin{table*}[t!]
\centering
\resizebox{0.95\textwidth}{!}{
\begin{tabular}{l|ccc|ccc|ccc}
\toprule
                  & \multicolumn{3}{c|}{\textbf{MuSiQue}}       & \multicolumn{3}{c|}{\textbf{2WikiMultiHopQA}} & \multicolumn{3}{c}{\textbf{HotpotQA}}       \\ \cmidrule{2-10}
                  & \textbf{EM}    & \textbf{F1}   & \textbf{Ratio}   & \textbf{EM}    & \textbf{F1}   & \textbf{Ratio}   & \textbf{EM}    & \textbf{F1}   & \textbf{Ratio}  \\ \midrule\midrule

\rowcolor[HTML]{EFEFEF}
\textit{Oracle}     & 21.53  & 37.51 & 0.15     & 36.6 & 51.3 & 0.42 & 41.40 & 58.95 & 0.17     \\
\textit{Raw Documents}     & 9.69  & 19.92 & (1.0)     & 24.06  & 36.75 & (1.0)      & 31.60 & 43.63 & (1.0)     \\
\midrule
Selective-Context & 3.02  & 9.96  & 0.15  & 18.87  & 29.30 & 0.36   & 17.74 & 28.82 & 0.17  \\ 
RECOMP (extractive)            & 8.32  & 16.76  & 0.16  & 22.26  & 33.89 & 0.45   & 28.09  & 40.78 & 0.19  \\ 
LLMLingua & 2.61  & 8.96  & 0.15  & 18.02  & 26.58 & 0.43   & 18.54 & 28.66 & 0.21  \\ 
LLMLingua-2 & 6.62  & 14.72  & 0.17  & 18.70  & 33.35 & 0.46   & 24.01 & 37.76 & 0.19 \\
LongLLMLingua & 10.95  & 19.66  & 0.19  & \underline{23.19} & \underline{34.83} & 0.50  & 31.68 & 46.49 & 0.26  \\
R2C & \underline{12.29}  & \underline{22.44}  & 0.16 & 19.41 & 32.09 & 0.36  & \underline{33.63} & \underline{47.80} & 0.22  \\
\midrule
IterCOMP (ours)   & \textbf{16.67} & \textbf{27.36} & 0.14  & \textbf{27.39} & \textbf{39.69} & 0.37   & \textbf{37.65} & \textbf{51.78} & 0.19  \\
\bottomrule
\end{tabular}
}
\caption{Experimental results on three multi-hop QA benchmark datasets. \textbf{Ratio} denotes the compression ratio, defined as the proportion of tokens in the compressed prompt relative to the those in the raw documents. The best performance is highlighted in \textbf{bold}, and the second-best is highlighted with an \underline{underline}. The full experimental results are presented in Appendix~\ref{appendix:exp_full}.}
\label{Tab:qa}
\end{table*}

\subsubsection{Answerability Judgment \& Missing Information Identification} 
The answerability judgment module serves as a reasoning controller, establishing a dynamic feedback loop for iterative evidence accumulation. At iteration $h$, the currently accumulated candidate evidence set is denoted as $E_\text{cand}^{(h)}$. The module employs an LLM as a binary classifier to determine whether $E_\text{cand}^{(h)}$ contains sufficient information to answer the original question $q^{(0)}$.

If the judgment is \textit{answerable}, the framework triggers an \textit{early termination}. The evidence set is finalized and used as the compressed prompt, $P_\text{comp}=E_\text{cand}^{(h)}$, and passed to the reader model $M$ to generate the final answer. This principled stopping condition prevents unnecessary iterations once sufficient information has been gathered. Conversely, if the judgment is \textit{unanswerable}, the framework proceeds with \textit{iterative refinement}. The LLM explicitly identifies the missing information that prevents a complete answer to the original question $q^{(0)}$. To bridge the gap, it formulates a targeted follow-up question $q^{(h)}$ to retrieve complementary evidence. Relevant evidence filtering is subsequently re-applied with respect to $q^{(h)}$, producing an updated set $E_\text{cand}^{(h+1)}$ progressively accumulated across iterations. This cycle repeats until the sufficiency condition is satisfied or a predefined maximum hop limit is reached.

\section{Experiments}

\subsection{Experimental Setup}

We evaluate our proposed prompt compression method, IterCOMP, in a zero-shot setting on three widely used multi-hop QA datasets: MuSiQue~\cite{trivedi-etal-2022-musique}, 2WikiMultiHopQA~\cite{ho-etal-2020-constructing}, and HotpotQA~\cite{yang-etal-2018-hotpotqa}. All experiments are conducted on dev sets of these datasets. For MuSiQue, we specifically use the \texttt{musique\_ans\_v1.0\_dev} subset, which contains only answerable questions. Further detailed statistics and descriptions of the datasets are provided in the Appendix~\ref{appendix:dataset}.

We compare IterCOMP against several baselines. To ensure a fair and meaningful comparison, we focus our evaluation on hard prompt compression and extractive compression strategies, which are most relevant to our approach. These include representative prompt compression baselines such as LLMLingua~\cite{jiang-etal-2023-llmlingua}, LongLLMLingua~\cite{jiang-etal-2024-longllmlingua}, LLMLingua-2~\cite{pan-etal-2024-llmlingua}, RECOMP~\cite{xu2024recomp}, Selective-Context~\cite{li-etal-2023-compressing}, and R2C~\cite{choi-etal-2024-reading}. We additionally report the performance of \textit{Oracle}, which serves as an upper bound for document-level compression by providing the reader model only with the gold supporting documents. In contrast, \textit{Raw Document} concatenates all documents in datasets without any filtering, representing a no-compression scenario. Detailed explanations of these baselines are available in Appendix~\ref{appendix:baseline}.

We employ \texttt{LLaMA-3-8B}~\cite{dubey2024llama} as the reader model across all evaluated methods and baselines. For IterCOMP, we bound the maximum number of iterations at 5 and set the hyperparameter $\lambda$, which balances the dual-aspect relevance score, to 0.6. The percentile threshold $k$ for relevance filtering is empirically determined and aligned with the \textit{Oracle} compression setting for fair comparison: $k=90$ for MuSiQue and HotpotQA, and $k=85$ for 2WikiMultiHopQA.

\subsection{Experimental Results}

\paragraph{Main Results.} To evaluate the effectiveness of our proposed IterCOMP, we employ Exact Match (EM) and F1 scores as metrics for QA performance, alongside the compression ratio (Ratio) to measure efficiency gains. As shown in Table~\ref{Tab:qa}, IterCOMP consistently achieves the highest EM and F1 scores across all benchmark datasets, significantly outperforming the baselines. Notably, it yields substantial improvements over the \textit{Raw Documents} setting, in which no compression is applied. For instance, on the MuSiQue dataset, the F1 score increases from 19.92 to 27.36, while on HotpotQA it rises from 43.63 to 51.78. These results demonstrate that our proposed adaptive compression framework enhances downstream QA performance while reducing input length (e.g., a 7$\times$ reduction on MuSiQue) by effectively filtering out irrelevant content from lengthy raw documents.

\begin{table}[t!]
\centering 
\resizebox{0.9\columnwidth}{!}{
\begin{tabular}{c|cc|cc}
\toprule
      & \textbf{EM} & \textbf{F1}     & \textbf{\# Iteration} & \textbf{\# Tokens}  \\ \midrule\midrule
2-hop & 19.33 & 30.19 & 1.98 & 228   \\ 
3-hop & 14.87 & 26.16 & 2.45 & 348  \\ 
4-hop & 11.60 & 20.72 & 3.08 & 403  \\ \bottomrule
\end{tabular}
}
\caption{Experimental results across different hop lengths. \textbf{\# Iterations} indicates the average number of iteration loops per question, and \textbf{\# Tokens} denotes the token lengths of the compressed prompt.}
\label{Tab:musique}
\end{table}

Compared to existing compression methods, IterCOMP demonstrates notable performance gains. On MuSiQue, it attains an F1 score of 27.36, surpassing the second-best method, R2C, by 4.92 points. Importantly, these results are achieved in a training-free manner, underscoring the generalizability and practicality of our approach. The proposed framework is designed to bridge missing-information gaps in multi-hop QA reasoning. This iterative refinement design dynamically retains only the most relevant information and ensures high performance even under strict input length constraints.

To further contextualize the results, we compare performance against the \textit{Oracle} setting, which serves as an upper bound based on ideal document selection. On HotpotQA, the performance gap between \textit{Raw Documents} and \textit{Oracle} is 15.32 points. IterCOMP attains an F1 score of 51.79, effectively closing 53.2\% of this gap. These results highlight the effectiveness of our framework in identifying and retaining compact yet highly informative evidence that supports complex multi-hop reasoning.

\noindent
\paragraph{Reasoning Complexity.} To assess the robustness of IterCOMP under varying reasoning complexity, we evaluate its performance on the MuSiQue dataset with respect to hop length. As shown in Table~\ref{Tab:musique}, performance declines as reasoning paths become longer. Specifically, the F1 score drops from 30.19 at 2-hop to 20.72 at 4-hop. This decline trend reflects the inherent difficulty of maintaining a coherent evidence chain and further indicates that increasing reasoning complexity imposes additional challenges on the reader model. 

The average number of iterations also increases with hop length, rising from 1.98 for 2-hop to 3.08 for 4-hop, indicating that IterCOMP dynamically allocates additional refinement steps when confronted with more complex questions. Similarly, the length of the compressed prompt expands from 228 to 403 tokens, reflecting that the framework adaptively retains a larger evidence base to support longer reasoning chains. Overall, rather than operating as a static filter, IterCOMP adjusts both its reasoning depth and evidence size in accordance with task complexity, thereby facilitating more effective handling of challenging multi-hop reasoning scenarios.

\subsection{Ablation Study \& Analysis}
As shown in Table~\ref{Tab:ablation}, we conduct an ablation study on the MuSiQue dataset to assess the contribution of each component in IterCOMP. Removing \textit{Iterative Refinement} leads to the largest performance drop, confirming the necessity of evidence accumulation in multi-hop reasoning. Excluding \textit{Relevant Evidence Filtering} (no-compression) also results in poor performance (19.92 F1), underscoring the importance of discarding irrelevant context. The absence of the \textit{Answerability Judgment} module degrades performance to 23.63 F1 and produces less compact prompts, validating the role of early stopping for both accuracy and efficiency. For the similarity measure, relying solely on semantic or lexical signals is suboptimal, whereas their combination achieves the best performance, demonstrating the synergistic complementarity of semantic and lexical cues.

\begin{table}[t]
\centering
\resizebox{1.0\columnwidth}{!}{
\begin{tabular}{lccc}
\toprule
 & \textbf{EM} & \textbf{F1} & \textbf{Ratio} \\
\midrule\midrule
- & \textbf{16.67} & \textbf{27.36} & \textbf{0.14} \\ \hline 
\rowcolor[HTML]{EFEFEF}\multicolumn{4}{l} {\textit{(without)}}\\
\quad \textit{Relevant Evidence Filtering}                  & 9.69 & 19.92 & (1.0) \\
\quad \textit{Answerability Judgment} & 11.67 & 23.63 & 0.21 \\
\quad \textit{Iterative Refinement}    & 8.50 & 17.39 & 0.06 \\ 
\cmidrule(lr){1-4}
Semantic Similarity Score  & 15.91 & 27.03 & 0.21 \\
Lexical Similarity Score & 14.57 & 25.19 & 0.27 \\
\bottomrule
\end{tabular}
}
\caption{Ablation study of IterCOMP, evaluating the role of each component and comparing different similarity measures for evidence filtering.}
\label{Tab:ablation}
\end{table}

\begin{figure}[t!]
  \centering
  \includegraphics[width=1.0\linewidth]{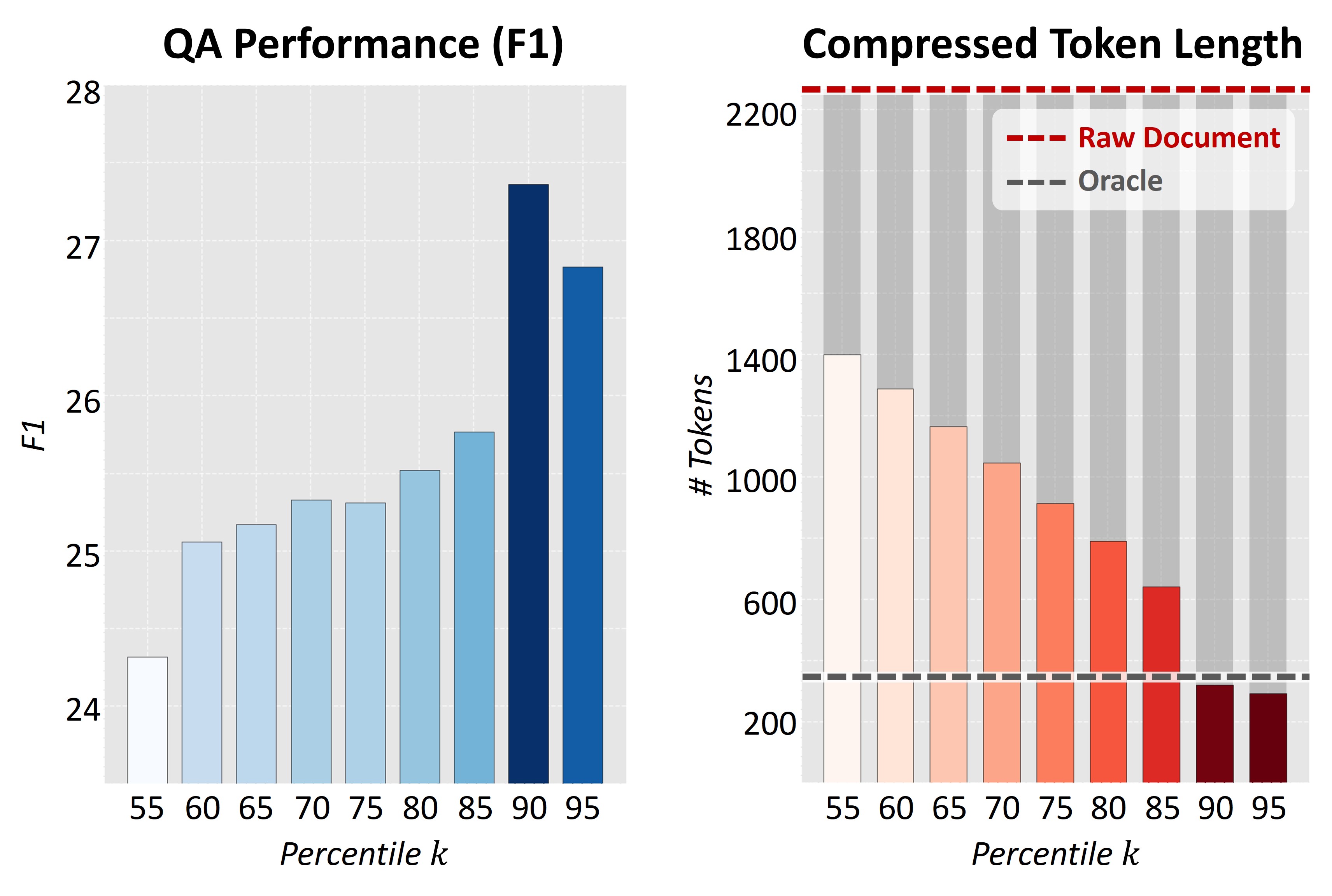}
  \caption{Comparison of \textit{(left)} QA performance and \textit{(right)} compressed token length across different percentile values $k$.}
  \label{percentile_fig}
\end{figure}

We evaluate different percentile values $k$ for relevant evidence filtering on the MuSiQue dataset, as shown in Figure~\ref{percentile_fig}. \textit{(Left)} The F1 score increases steadily with larger $k$, peaking at 27.36 when $k=90$. This indicates that a stricter relevance threshold effectively isolates salient information and reduces noise, thereby enhancing the reader model's reasoning. However, performance drops at $k=95$, suggesting that an overly stringent cutoff discards supplementary evidence necessary for completing the reasoning chain. \textit{(Right)} As $k$ increases, compressed prompts become shorter due to more aggressive filtering. Notably, when $k \geq 90$, the compressed prompt is even more concise than that of the \textit{Oracle}.

\begin{table}[t!]
\resizebox{1.0\columnwidth}{!}{
\begin{tabular}{lcc}
\toprule
             & \textbf{Cost Reduction (\%)} & \textbf{Speedup} \\ \midrule\midrule
\texttt{GPT-3.5-Turbo} & 79.4  & 1.13$\times$ \\ 
\texttt{GPT-4o}   & 75.5 & 1.19$\times$ \\ 
\texttt{Gemini-2.5-Flash}   & 76.5  & 1.40$\times$   \\ 
\texttt{Gemini-2.5-Pro}      & 79.4          & 1.87$\times$      \\ 
\texttt{Claude-4.5}    & 76.0   & 2.01$\times$ \\ \bottomrule
\end{tabular}
}
\caption{Efficiency analysis with API-based LLMs. The reported cost and latency correspond to the final QA step performed by the reader LLM with the compressed prompt.}
\label{Tab:api}
\end{table}

To assess the practical utility and economic viability of our method, we analyze 500 random instances from the MuSiQue dataset using prominent API-based LLMs. As shown in Table~\ref{Tab:api}, applying our framework for prompt compression substantially improves inference efficiency. Across all models, compression reduces costs by 75.5\%--79.4\%, making multi-hop QA with large-scale commercial LLMs considerably more affordable. In addition, compressed prompts consistently accelerate inference, highlighting prompt compression as a practical enabler of scalable, time-efficient, and cost-effective deployment of advanced complex reasoning in commercial LLMs.

\section{Conclusion}

This paper introduces IterCOMP, a unified prompt compression framework for multi-hop QA. IterCOMP embeds multi-step reasoning into the compression loop through iterative refinement, combining answerability judgment with targeted follow-up question generation. This process progressively constructs a concise, information-rich prompt by retaining only the evidence essential for complex reasoning. Extensive experiments on three multi-hop QA benchmarks demonstrate that IterCOMP significantly enhances QA performance while reducing token usage. Its training-free and model-agnostic design ensures broad applicability, including integration with commercial black-box LLMs for which fine-tuning is infeasible. By addressing both the performance degradation in long contexts and the economic burden of large-scale inference, IterCOMP offers a robust and scalable solution for complex multi-document LLM applications.

\section*{Limitations}

While IterCOMP demonstrates improvements in multi-hop QA, several aspects remain open for refinement. First, its effectiveness depends on the reasoning capability of the underlying LLM for answerability judgment and missing information detection, introducing the risk of error propagation such as premature termination or accumulation of irrelevant evidence. Second, the iterative design incurs additional compression overhead compared to single-pass methods, which could be alleviated by lightweight controllers or parallel filtering. Third, performance is sensitive to hyperparameters such as the relevance percentile and iteration limit; adaptive mechanisms that dynamically adjust these settings based on question complexity or intermediate retrieval quality could further enhance robustness. Fourth, our lexical-semantic scoring module adopts a relatively simple combination of signals, and integrating more sophisticated scoring schemes may yield additional performance gains. Finally, our evaluations focus on general-domain multi-hop QA benchmarks to ensure fair comparison with existing baselines; although IterCOMP is modular and domain-agnostic by design, validating its generalization to diverse reasoning types and non-Wikipedia domains remains an important direction for future work.

\section*{Acknowledgments}
This work was supported by the Institute of Information \& Communications Technology Planning \& Evaluation (IITP) grant funded by the Korea government (MSIT) [RS-2021-II211341, Artificial Intelligence Graduate School Program (Chung-Ang
University)] and by the National Research Foundation of Korea (NRF) grant funded by the Korea
government (MSIT) (RS-2025-00556246).


\bibliography{custom}

\newpage 

\appendix

\newpage

\section{Prompts}

\begin{table}[h!]
\resizebox{1.0\columnwidth}{!}{
\begin{tabular}{p{8cm}}
\toprule
\rowcolor[HTML]{EFEFEF} 
\textbf{Prompt for Sub-QA Pair Generation} \\ 
Given a question, its final answer, and evidences, please generate a sub-question and sub-answer for each evidence. Each pair must be reflect the content of its corresponding evidence. \\
\\
Question: When was the creator of The Painter's Studio born? \\
Answer: 10 June 1819 \\
Evidence: [{"question": "The Painter's Studio >> creator", "answer": "Gustave Courbet"}, {"question":
"The date of birth of \#1 is?", "answer": "10 June 1819"}] \\
\\
\textit{(generated sub-QA example)}\\
Sub-question 1: Who is the creator of The Painter's Studio? \\
Sub-answer 1: The creator of The Painter's Studio is Gustave Courbet. \\
Sub-question 2: What is the date of birth of Gustave Courbet? \\
Sub-answer 2: The date of birth of Gustave Courbet is 10 June 1819. \\
\midrule
\rowcolor[HTML]{EFEFEF} 
\textbf{Answerability Judgment \& Missing Information Identification} \\ 
Given the following Question and Information, your task is to determine whether the Information alone is sufficient to answer the Question.\\
\\
If the Information provides a clear and direct answer, of if answer to the Question can be logically derived by combining multiple statements within the Information, then output: "answerable".
\\ \\
If the Information lacks the necessary content to fully answer the Question, then output: "unanswerable", and generate a follow-up question that identifies the most specific and essential piece of missing information required to answer the Question. \\ \\
Your response must strictly follow this format: ["answer": "answerable or unanswerable", "follow\_up\_question": "a specific and detailed question that would help retrieve the missing information"] \\ \\
Do not include any explanation or reasoning outside of this format.
\\
\bottomrule
\caption{Prompts and generated examples used for sub-QA pair generation (top) and answerability judgment with missing information identification (bottom).}
\label{tab:analysis}
\end{tabular}
}
\end{table}

\section{Implementation Details}
\label{appendix:B}
\subsection{Datasets}
\label{appendix:dataset}

\begin{itemize}
    \item \textbf{MuSiQue}~\cite{trivedi-etal-2022-musique} consists of questions constructed by combining multiple single-hop queries, requiring 2 to 4 reasoning hops, and is specifically designed to prevent answer derivation from superficial clues. 
    \item \textbf{2WikiMultiHopQA}~\cite{ho-etal-2020-constructing} is a Wikipedia-based dataset featuring questions that require up to five reasoning steps, with each question annotated with its explicit reasoning path and supporting evidence.
    \item \textbf{HotpotQA}~\cite{yang-etal-2018-hotpotqa} requires models to retrieve and synthesize evidence from multiple documents, containing explainable questions explicitly designed to encourage transparent and interpretable reasoning paths.
\end{itemize}
\noindent
Table~\ref{Tab:statistic} presents the statistics of the benchmark multi-hop QA datasets used in our experiments. We employ three widely used standard datasets: MusiQue, 2WikiMultiHopQA, and HotpotQA. For each dataset, we summarize the number of instances in the train, dev, and test splits.

\begin{table}[h!]
\centering
\resizebox{1.0\columnwidth}{!}{
\begin{tabular}{l|ccc|c}
\toprule 
& \textbf{Train}    & \textbf{Dev}  & \textbf{Test} & \textbf{\# Context} \\ \midrule\midrule
MuSiQue & 39876  & 4834 & 4918 & 20   \\ 
(MuSiQue-Ans) & (19938) & 2417 & (2459) & (20)   \\ \addlinespace
2WikiMultiHopQA & 167454 & 12576 & 12576 & 10   \\ \addlinespace
HotpotQA & 90447 & 7405 & - & 10 \\ \bottomrule
\end{tabular}
}
\caption{Statistics of multi-hop QA datasets used in our experiments.}
\label{Tab:statistic}
\end{table}

\begin{table*}[t!]
\centering
\resizebox{0.85\textwidth}{!}{
\begin{tabular}{l|ccc|ccc|ccc}
\toprule
                  & \multicolumn{3}{c|}{\textbf{MuSiQue}}       & \multicolumn{3}{c|}{\textbf{2WikiMultiHopQA}} & \multicolumn{3}{c}{\textbf{HotpotQA}}       \\ \cmidrule{2-10}
                  & \textbf{EM}    & \textbf{F1}   & \textbf{Ratio}   & \textbf{EM}    & \textbf{F1}   & \textbf{Ratio}   & \textbf{EM}    & \textbf{F1}   & \textbf{Ratio}  \\ \midrule\midrule

\rowcolor[HTML]{EFEFEF}
\textit{Oracle}     & 21.53  & 37.51 & 0.15     & 36.6 & 51.3 & 0.42 & 41.40 & 58.95 & 0.17     \\
\textit{Raw Documents}     & 9.69  & 19.92 & (1.0)     & 24.06  & 36.75 & (1.0)      & 31.60 & 43.63 & (1.0)     \\
\midrule
Selective-Context & 3.02  & 9.96  & 0.15  & 18.87  & 29.30 & 0.36   & 17.74 & 28.82 & 0.17  \\ 
RECOMP (extractive)            & 8.32  & 16.76  & 0.16  & 22.26  & 33.89 & 0.45   & 28.09  & 40.78 & 0.19  \\ 
RECOMP (abstractive)            & 4.18  & 11.57  & 0.03 & 13.34  & 33.08 & 0.07  & 21.74  & 38.44 & 0.05  \\ 
LLMLingua & 2.61  & 8.96  & 0.15  & 18.02  & 26.58 & 0.43   & 18.54 & 28.66 & 0.21  \\ 
LLMLingua-2 & 6.62  & 14.72  & 0.17  & 18.70  & 33.35 & 0.46   & 24.01 & 37.76 & 0.19 \\
LongLLMLingua & 10.95  & 19.66  & 0.19  & \underline{23.19} & \underline{34.83} & 0.50  & 31.68 & 46.49 & 0.26  \\
CompAct & 8.27  & 18.77  & 0.07 & 13.32 & 33.1 & 0.16  & 26.12 & 42.98 & 0.12  \\
R2C & \underline{12.29}  & \underline{22.44}  & 0.16 & 19.41 & 32.09 & 0.36  & \underline{33.63} & \underline{47.80} & 0.22  \\
\midrule
IterCOMP (ours)   & \textbf{16.67} & \textbf{27.36} & 0.14  & \textbf{27.39} & \textbf{39.69} & 0.37   & \textbf{37.65} & \textbf{51.78} & 0.19  \\
\bottomrule
\end{tabular}
}
\caption{Full experimental results on three multi-hop QA benchmarks, 
including abstractive RECOMP and CompAct.}
\label{Tab:full}
\end{table*}

\subsection{Baselines}
\label{appendix:baseline}

To ensure reproducibility, all experiments were conducted using officially released codebases and publicly available models.
\begin{itemize}
    \item For \textbf{RECOMP}~\cite{xu2024recomp}, we implement an extractive compressor based on a dual-encoder model that identifies and selects relevant sentences, representing a sentence-level extractive strategy.
    \item \textbf{Selective-Context}~\cite{li-etal-2023-compressing} is a token-level compression approach that removes low-information lexical units using a compact language model.
    \item \textbf{LLMLingua}~\cite{jiang-etal-2023-llmlingua} leverages a smaller language model to eliminate low-perplexity tokens from the original prompt to satisfy a predefined compression ratio.
    \item \textbf{LLMLingua-2}~\cite{pan-etal-2024-llmlingua} is an extension of LLMLingua that incorporates a refined budget controller and data distillation mechanisms to more effectively preserve essential information.
    \item \textbf{LongLLMLingua}~\cite{jiang-etal-2024-longllmlingua} extends LLMLingua, introducing a question-aware, coarse-to-fine compression strategy to retain critical information from lengthy documents.
    \item \textbf{R2C}~\cite{choi-etal-2024-reading} compresses prompts by leveraging cross-attention scores from a Fusion-in-Decoder (FiD) model to score the relevance of each chunk and sentence, retaining only those identified as most important.
\end{itemize}
\noindent

\subsection{Additional Experiments Results}
\label{appendix:exp_full}

RECOMP~\cite{xu2024recomp} also supports abstractive compression, which generates summaries by synthesizing information across retrieved documents. Moreover, CompAct~\cite{yoon-etal-2024-compact} compresses retrieved documents by jointly analyzing previously selected content alongside newly introduced segments, enabling context-aware compression across the document set. 

However, a fair comparison with these methods is challenging due to inconsistencies in length control. While neither our method nor most baselines enforce strict token-level constraints, both abstractive RECOMP and CompAct exhibit substantially larger deviations from the target token budget. These methods rely on free-form generation without an explicit length-control mechanism and fail to reliably regulate the length of the compressed prompt. Consequently, we exclude these methods from the main comparison and instead report their full results in Table~\ref{Tab:full}, explicitly noting the fairness limitations arising from these inconsistent compression budgets. As shown in Table~\ref{Tab:full}, IterCOMP still consistently outperforms all baselines, including abstractive RECOMP and CompAct, across the three benchmarks.

\end{document}